\pdfoutput=1

\documentclass[11pt]{article}

\usepackage{naacl2021}

\usepackage{times}
\usepackage{latexsym}
\usepackage{multirow}
\usepackage{graphicx}

\usepackage[T1]{fontenc}

\usepackage[utf8]{inputenc}

\usepackage{microtype}

\title{{\sc MUSER:} \underline{MU}ltimodal \underline{S}tress Detection using \\\underline{E}motion \underline{R}ecognition as an Auxiliary Task}

\author{\normalsize Yiqun Yao$^\textrm{{1}}$, {Michalis Papakostas}$^\textrm{{1}}$, Mihai Burzo$^\textrm{{2}}$, Mohamed Abouelenien$^\textrm{{3}}$, Rada Mihalcea$^\textrm{{1}}$ \\
  $^\textrm{1}$Computer Science and Engineering, University of Michigan \\
  $^\textrm{2}$Mechanical Engineering, University of Michigan \\
  $^\textrm{3}$Computer and Information Science, University of Michigan \\
  {\tt \{yaoyq,mpapakos,mburzo,zmohamed,mihalcea\}@umich.edu} \\}

\date{}

\begin{document}
\maketitle
\begin{abstract}
The capability to automatically detect human stress can benefit artificial intelligent agents involved in affective computing and human-computer interaction. Stress and emotion are both human affective states, and stress has proven to have important implications on the regulation and expression of emotion. Although a series of methods have been established for multimodal stress detection, limited steps have been taken to explore the underlying inter-dependence between stress and emotion. In this work, we investigate the value of emotion recognition as an auxiliary task to improve stress detection. We propose {\sc MUSER} -- a transformer-based model architecture and a novel multi-task learning algorithm with speed-based dynamic sampling strategy. Evaluations on the Multimodal Stressed Emotion (MuSE) dataset show that our model is effective for stress detection with both internal and external auxiliary tasks, and achieves state-of-the-art results.
\end{abstract}

\section{Introduction}
Stress is a feeling of emotional or physical tension, as a response to the environment when people have difficulty dealing with the conditions \cite{dobson2000stress, muthukumar2013phosphatidylethanolamine}. Stress detection is a classification task that predicts whether a certain target is under stress. The task has drawn research attention for two reasons: first, stress detection plays an important role in applications related to psychological well-being \cite{cohen1991psychological}, cognitive behavior therapies \cite{tull2007preliminary}, and safe driving \cite{gao2014detecting,chen2017detecting}; second, stress is a known regulator of human emotion mechanisms \cite{tull2007preliminary}, and thus research on stress detection can potentially benefit the development of emotionally intelligent agents.

The impact of stress on human behavior can be observed through various modalities. Previous work has considered both unimodal and multimodal stress detection using acoustic, video and physiological sensor signals \cite{lane2015deepear,jaques2016multi,aigrain2016multimodal,alberdi2016towards,bara2020deep}. However, text-based stress detection remains vastly underexplored, with some studies \cite{lin2014user} showing the potential for further research. In recent years, the surge of advanced natural language understanding models and structures provides a great opportunity for stress detection systems, especially using the textual modality. In this work, we focus on the textual and acoustic modalities. For the model architecture, we use Transformers \cite{vaswani2017attention} as a textual encoder and Multi-Layer Perceptrons (MLP) as an acoustic encoder.

The majority of existing stress detection methods are based on single-task learning with the binary stress/non-stress labels. However, stress is not an isolated affective state, but closely related to the expression and regulation of human emotions. Physiological studies \cite{wang2011emotion} have demonstrated that emotion and stress share some neural structures, including prefrontal cortex \cite{taylor2007coping}, anterior cingulate cortex \cite{pruessner2008deactivation}, and amygdala \cite{adolphs2003human}. Acoustic studies  \cite{paulmann2016psychological}  have shown that the pitch and amplitude of human emotional prosody is different under stress and non-stressed status. Inspired by these studies, our work aims to exploit the inter-dependence between emotion and stress. Specifically, we investigate the value of emotion recognition as an auxiliary task for stress detection.

Multi-task learning \cite{pasunuru2017multi,gottumukkala2020dynamic,guo2018dynamic,gong2019comparison} has proven to be effective for transferring knowledge between different tasks. Dynamic sampling strategies, which aim at adaptively adjusting the ratio of samples from different tasks, are widely used to balance the training schedule. However, strategies based on gradients \cite{chen2018gradnorm}, uncertainty \cite{kendall2018multi} or loss \cite{liu2019end} cannot leverage the validation performances, while some performance-based strategies \cite{gottumukkala2020dynamic} are impractical if the metrics for different tasks are not directly comparable (i.e., with different scale ranges). To this end, we propose a novel speed-based strategy that is both effective and efficient in the multi-task learning for stress and emotion.

Our method is evaluated on the Multimodal Stressed Emotion (MuSE) dataset \cite{jaiswal2019muse,jaiswal2020muse}, which includes  both stress and emotion labels, making it the ideal  benchmark for an in-depth analysis of their inter-dependence. To test the generalization ability of our method, we also use an external emotion dataset for the auxiliary task. Multimodal emotion recognition is a well-studied field with many existing datasets \cite{busso2008iemocap,busso2016msp,chen2018emotionlines,barros2018omg,zadeh2018multimodal}. We choose the OMG-Emotion dataset \cite{barros2018omg} as the external auxiliary task because it is representative and challenging, with numerical emotion scores instead of categorical labels.

Our paper makes four main contributions. First, we show the inter-dependence between stress and emotion via quantitative analyses on linguistic and acoustic features, and propose to use emotion recognition as an auxiliary task for stress detection. Second, we establish a stress detection model with a transformer structure, as well as a novel speed-based dynamic sampling strategy for multi-task learning. We name our framework the \underline{MU}ltimodal \underline{S}tress Detector with \underline{E}motion \underline{R}ecognition ({\sc MUSER}). Third, we achieve state-of-the-art results on the MuSE dataset via multi-task training with stress and emotion labels. We also achieve competitive results  when we  use the OMG-Emotion \cite{barros2018omg} dataset as an external auxiliary task. Finally, experimental results show that our speed-based dynamic sampling significantly outperforms other widely-used methods.

\section{Related Work}
\subsection{Unimodal Stress Detection}
Stress detection based on textual modality has been studied by \cite{lin2014user} and \cite{jaiswal2020muse}, using the Linguistic Inquiry and Word Count (LIWC) lexicon \cite{pennebaker2001linguistic} to extract features that are indicative of human emotion. Acoustic features \cite{lane2015deepear,paulmann2016psychological,horvath1982detecting,lech2014stress} have also been used for unimodal stress detection in both physiological and computational studies.

A drawback of the unimodal approaches is that they only have access to partial information about the expression of stress, while multiple modalities can potentially be informative at the same time \cite{aigrain2016multimodal}. As demonstrated by previous work on human sentiment and emotion prediction \cite{zadeh2016mosi,zadeh2018multimodal,yao2020morse}, multimodal features usually results in better performances.

\subsection{Multimodal Stress Detection}
Commonly-used modalities for stress detection include video, audio, text and physiological signals such as thermal maps from sensors \cite{aigrain2016multimodal,alberdi2016towards,lane2015deepear,jaques2016multi}.

Jaiswal et al. \shortcite{jaiswal2020muse} proposed the Multimodal Stressed Emotion (MuSE) dataset, which includes records from all the commonly-used modalities. Each video clip is annotated for both stress detection and emotion recognition. Unimodal and multimodal baselines are provided for each task. Bara et al. \shortcite{bara2020deep} developed a multimodal deep learning method that learns modality-independent representations in an unsupervised approach. However, none of these models leverage the intrinsic connections between stress and emotion. 

Our experiments are conducted on the MuSE dataset using only the textual and acoustic modalities, to be compatible with most external emotion recognition tasks. However, our proposed multi-task learning method is model-agnostic and can be generalized to any structure and any modality combinations.

\subsection{Emotion Recognition}
Widely-used multimodal emotion recognition datasets include SEMAINE \cite{mckeown2011semaine}, IEMOCAP \cite{busso2008iemocap}, MOSEI \cite{zadeh2018multimodal} and OMG-Emotion \cite{barros2018omg}. Emotion can be annotated either with pre-defined emotion categories or through two-dimensional scores of activation (arousal) and valence, according to the self-assessment manikin proposed by \cite{bradley1994measuring}. MuSE, in particular, has emotion annotations expressed by activation and valence scores (1$\sim$9), which is more fine-grained than categorical definitions (happy, angry, etc.). The OMG-Emotion dataset we use as external auxiliary task is annotated in the same way with a score range of 0$\sim$1.

    

    

\subsection{Multi-task Learning}
Because of the different task natures, balancing the training procedure with all the tasks is a critical problem for multi-task learning. Loss-balancing strategies \cite{chen2018gradnorm,kendall2018multi,liu2019end,gong2019comparison,guo2018dynamic,lample2017unsupervised,yao2019world} are suitable for situations in which there are multiple training objectives that can be combined via weighted summation for each data point. In contrast, for multi-task learning across different datasets, a sampling strategy should be applied to decide the mixing ratio (how many batches to sample from each task) in each epoch. To this end, Pasunuru et al. \shortcite{pasunuru2017multi} used a fixed sampling ratio; Guo et al. \shortcite{feifei2018dynamic} proposed a dynamic sampling strategy based on reinforcement learning, which depends on the estimation of Q-values; Gottumukkala et al. \shortcite{gottumukkala2020dynamic} used a dynamic sampling procedure based on the gap between multi-task and single-task results -- a  performance-based method that requires all the tasks to use the same set of evaluation metrics. For comparison, our proposed strategy is also based on how fast the model is learning each task, but does not require the metrics to be directly comparable.

\begin{table}[t]
  \begin{center}
    \caption{LIWC features that have the top 20 highest regression coefficients in all three tasks.}
    \label{stress-language-common}
    \scalebox{0.9}{
    \begin{tabular}{l|l}
    \hline 
    Feature & Examples\\
    \hline
    bio & eat, blood, pain\\
    health & clinic, flu, pill \\
    relativity & area, stop, exit\\
    body & cheek, hands, spit\\
    ingest & dish, eat, pizza\\
    positive-emo & love, nice, sweet\\
    space & down, in, thin\\
    time & end, until, season\\
    perceptual & observe, hear, feeling\\\hline
    
    \end{tabular}
    }
  \end{center}
\end{table}

\section{Expressions of Stress in Data}
\subsection{Dataset}
The MuSE dataset \cite{jaiswal2020muse} is collected from the multimodal video recordings of 28 student participants, 9 female and 19 male. Each participant is invited to a video-recording session before and after the final exam period; sessions before exams are labeled as stressed, and the remainng ones are labeled as non-stressed. We use only the records from the monologue sub-sessions where both acoustic and textual modalities are available. In these sub-sessions, the participants view five emotion-eliciting questions on the screen in a sequence, and answer them with monologues. After each monologue, the participants provide self-assessment scores for activation (calm vs. excited) and valence (negative vs. positive). The scores range from 1$\sim$9. The monologues are segmented into sentences for pre-processing; each sentence is annotated with the same stress label and emotion scores as the whole monologue.

We use a train, validation, and test split of 1,853, 200, and 273 sentences, respectively. Textual features come from the automatic transcripts for the audio clips of each sentence. Although the sentences come with visual and thermal features as well, we focus mainly on the textual and acoustic modalities because this allows us to use almost any external emotion recognition dataset as our auxiliary task.

\begin{table}[t]
  \begin{center}
    \caption{LIWC features that are among the top 20 highest regression coefficients unique to  \textcolor{olive}{stress}, \textcolor{blue}{activation} and \textcolor{purple}{valence} tasks.}
    \label{stress-language}
    \scalebox{0.95}{
    \begin{tabular}{l|l}
    \hline 
    Feature & Examples\\
    \hline
    \textcolor{olive}{nonfl} & er, hmm, umm\\
    \textcolor{olive}{affect} & happy, cried, abandon\\
    \textcolor{olive}{social} & mate, talk, child\\
    \textcolor{olive}{family} & daughter, husband, aunt\\\hline
    \textcolor{blue}{past} & went, ran, had\\
    \textcolor{blue}{money} & audit, cash, owe\\
    \textcolor{blue}{tentat} & maybe, perhaps, guess\\\hline
    \textcolor{purple}{feel} & feels, touch\\
    \textcolor{purple}{sad} & crying, grief, sad\\
    \textcolor{purple}{negate} & no, not, never\\
    \textcolor{purple}{anger} & hate, kill, annoyed\\
    \textcolor{purple}{achieve} & earn, hero, win\\
    \textcolor{purple}{quant} & few, many, much\\\hline
    
    \end{tabular}
    }
  \end{center}
\end{table}

\begin{table*}[t]
  \begin{center}
    \caption{Opensmile eGeMap features that have the top 20 highest regression coefficients for different tasks.}
    \label{stress-speech}
    \scalebox{0.95}{
    \begin{tabular}{c|c|c|c}
    \hline 
    Task & Feature & Category & Examples\\
    \hline
    \multirow{3}{*}{All} & Amplitude & Energy & F1/F2/F3 mean amplitude\\
    & Loudness & Energy & mean loudness\\
    & Spectral Flux & Spectral & mean flux in voiced regions\\\hline
    \multirow{3}{*}{Only in \textcolor{olive}{stress}} & Hammarberg Index & Spectral & hammarberg index in unvoiced regions\\
    & Alpha Ratio & Spectral & mean ratio for unvoiced regions\\
    & Slope & Spectral & 500-1500Hz in unvoiced regions\\\hline
    \multirow{2}{*}{Only in \textcolor{blue}{activation}} & HNR & Energy & mean HNR (Harmonics-to-Noise Ratio)\\
    & Voice Length & Temporal & mean voiced segment length\\\hline
    \multirow{2}{*}{Only in \textcolor{purple}{valence}} & Pitch & Frequency & F0 semitone\\
    & Formant & Frequency & frequency of formant 3\\\hline
    
    \end{tabular}
    }
  \end{center}
\end{table*}

\subsection{Characteristics of Stress in Language}
\label{3.2}
In order to analyze the connections between linguistic features that are most indicative of stress, activation, and valence, we first extract a feature vector based on the LIWC lexicon \cite{pennebaker2001linguistic}. Each dimension of the vector corresponds to a certain word category and has a value equal to the count of words observed in that category in the sentence. We then apply z-normalization on each feature and fit a linear model to predict the stress/non-stress label, as well as the activation and valence scores. For each of the three tasks, we pick the features with the top 20 highest absolute values of the linear classification/regression coefficients, which we assume to be the key indicators. 

Features that appear in top 20 for all three tasks are shown in Table \ref{stress-language-common}. The features are ranked by the absolute value of their linear coefficients. As shown, the ``positive-emotion'' and ``perceptual'' word classes are critical for both emotion and stress tasks, which is intuitive because they are a pair of inter-dependent human affect status. Bio, health, and body words are also on the list, suggesting that both stress and emotion are closely related to physiological status and feelings, which is potentially because they share some neural structures in brain \cite{wang2011emotion}. The intersection of all the three top-indicator sets has nine elements, reflecting a reasonable overlap.

Table \ref{stress-language} shows the word classes appearing uniquely in the top 20 indicator list for each task. It is worth noticing that the non-fluent words (er, hmm, etc.) are the strongest unique indicator of stress, which reflects the differences in the audio speeches under stressed/non-stressed conditions. We could also observe that activation is more connected to entities and events, while valence is more related to personal feelings.

\subsection{Characteristics of Stress in Speech}
\label{3.3}
For stress indicators in the acoustic modality, we extract 88-dimensional features using OpenSmile \cite{eyben2010opensmile} with the eGeMaps \cite{eyben2015geneva} configuration. We follow \cite{jaiswal2020muse} to do speaker-level z-normalization on each feature, and fit a linear classification/regression model as we did for the textual features.

Table \ref{stress-speech} shows the most indicative acoustic feature classes for all the tasks, as well as the ones that are unique for each task. Amplitude/loudness is the strongest indicator class for all tasks, followed by spectral flux, which is a measure of how quickly the power spectrum of a signal is changing. It also suggests that stress has a closer relationship with spectral features such as slope, describing how quickly the spectrum of an audio sound tails off towards the high frequencies. 

The intersection of all three indicator sets has 11 elements, suggesting that they share many acoustic patterns. For more detailed explanations and examples of the eGeMaps features please refer to \cite{eyben2015geneva} and \cite{botelho2019speech}.

Regarding the differences in the task nature, as seen in Table \ref{uniq}, the number of unique indicators for each each and for each modality show that the activation task is less independent of the stress task than the valence task. In other words, the activation task has more indicators in common with the stress task.

\begin{table}[h]
  \begin{center}
    \caption{Number of unique textual and acoustic indicators for stress, activation and valence.}
    \label{uniq}
    \begin{tabular}{cccc}
    \hline 
    Feature & Stress & Activation & Valence\\\hline
    LIWC & 4 & 3 & 6\\
    eGeMaps & 7 & 6 & 8\\\hline
    
    \end{tabular}
  \end{center}
\end{table}

\begin{figure*}[h]
    \centering
    \includegraphics[width=16cm]{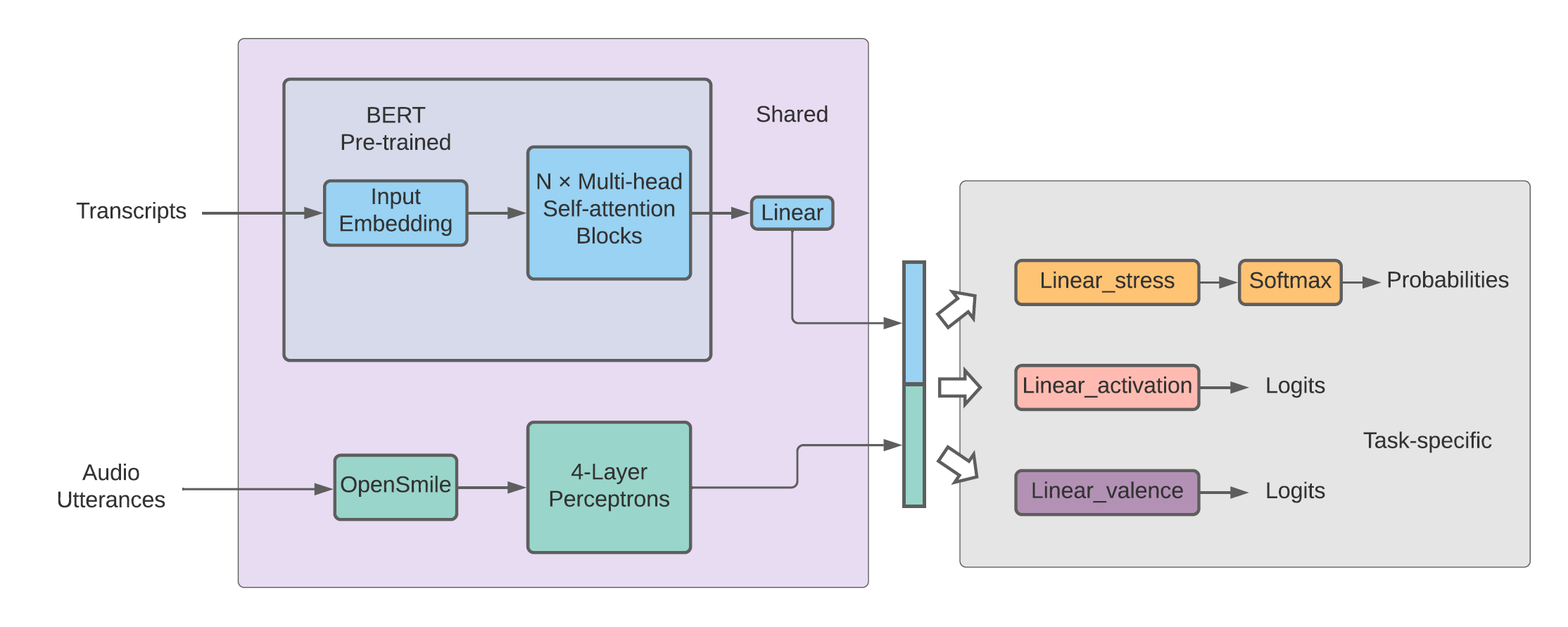}
    \caption{Multimodal fusion architecture for {\sc MUSER}.}
    \label{fig_late_fusion}
\end{figure*}

\section{Method}
\subsection{Auxiliary Tasks}
Based on the task inter-dependency demonstrated in Section \ref{3.2} and \ref{3.3}, we propose to use multimodal emotion recognition as an auxiliary task for stress detection. Since MuSE has both stress and emotion labels, their activation and valence scores can be used as an internal auxiliary task.

To test the generalization capability of our multi-task learning method, we choose OMG-Emotion \cite{barros2018omg} as an external emotion recognition dataset for the auxiliary task, which is annotated in the same manner as MuSE (activation/valence). We download the videos from the training and validation sets and filter out all the samples where the video link is broken or the length of automatic transcription is less than 5, resulting in 1,484 videos. The contents and scenarios in the OMG-Emotion dataset are completely different from MuSE. We hold out 300 videos as a validation set to enable dynamic sampling. 

Note that stress detection is a binary classification task, while the two auxiliary emotion tasks have a regressive nature.

\subsection{Pre-processing}
Each utterance in the MuSE dataset is automatically segmented into sentences, transcribed, and tokenized by a pre-trained BERT tokenizer \cite{devlin2019bert}. For the acoustic modality, we use OpenSmile \cite{eyben2010opensmile} with the eGeMAPS configuration \cite{eyben2015geneva} to extract 88 utterance-level statistical features. Following \cite{jaiswal2020muse}, we perform speaker-level z-normalization on all acoustic features. For videos in the OMG-Emotion dataset, we first extract the audio and automatic transcripts, and then do the same pre-processing as on MuSE.

\subsection{{\sc MUSER}: Architecture}
\label{4.3}
We propose {\sc MUSER}: \underline{MU}ltimodal \underline{S}tress Detector using \underline{E}motion \underline{R}ecognition. The model structure is based on neural networks. Specifically, we use a Transformer \cite{vaswani2017attention} textual encoder pre-trained with BERT \cite{devlin2019bert}, and an MLP-based acoustic encoder to generate representations on each modality, and fuse them before classification or regression. Our model architecture is depicted in Figure \ref{fig_late_fusion}.

\begin{figure*}[h]
    \centering
    \includegraphics[width=13cm]{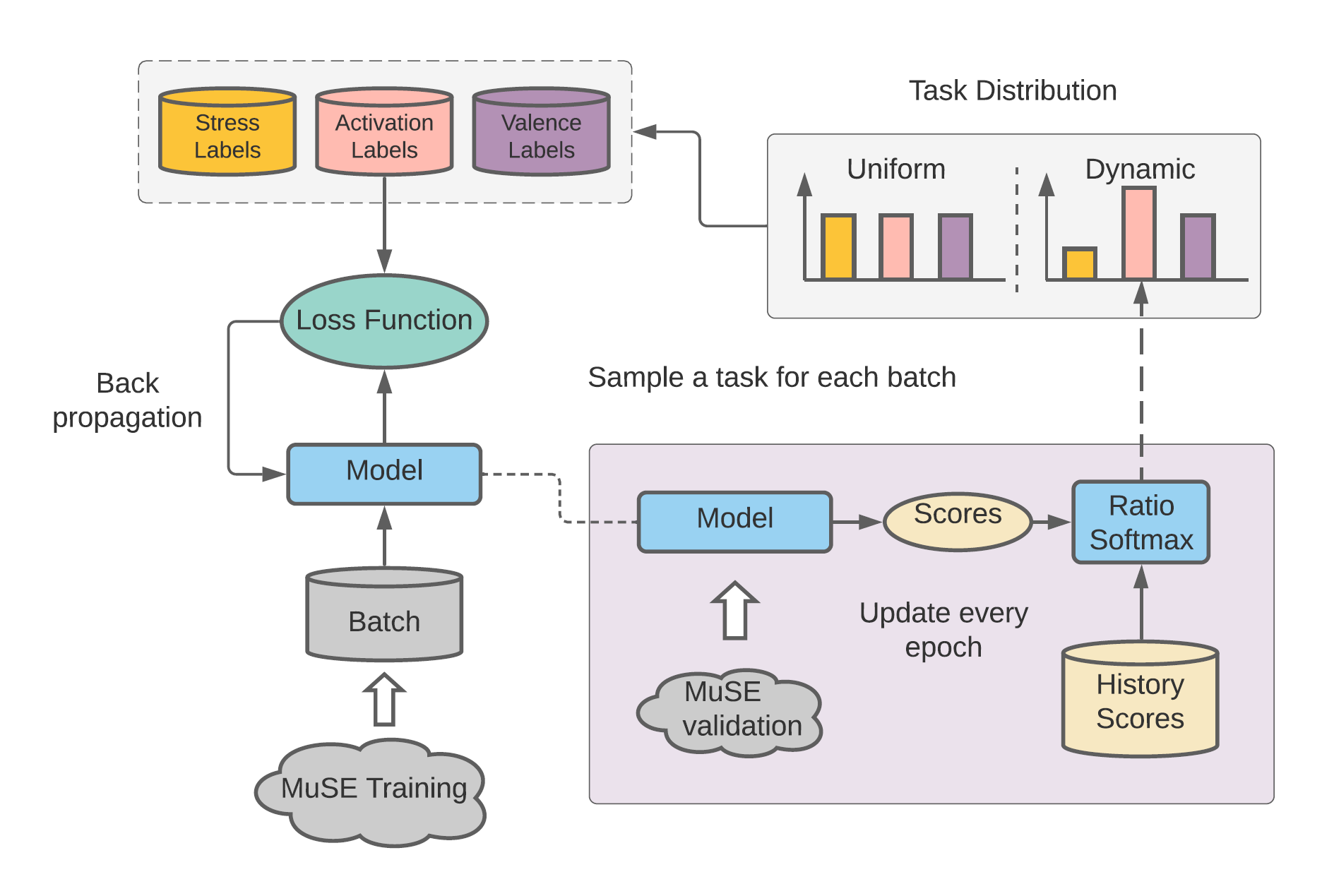}
    \caption{Dynamic sampling procedure for MUSER multi-task training.}
    \label{fig_dynamic}
\end{figure*}

\subsubsection{Textual Encoder}
For the textual encoder, we use a Transformer neural network pre-trained with BERT on BookCorpus and English Wikipedia \cite{devlin2019bert}. Our Transformer model has 12 layers, 12 attention heads, and 768-dimensional hidden states. The averaged hidden states on the top level are projected to 256-dimensional representations by a fully-connected layer.

\subsubsection{Acoustic Encoder}
Our acoustic encoder is a Multi-layer Perceptron network with four hidden layers and ReLU activation. The input of the acoustic encoder is the OpenSmile features extracted from the audio speech of each sentence, and the output of each hidden layer is 256-dimensional.

\subsubsection{Multimodal Fusion}
We fuse the multimodal features by concatenating the top-level 256-dimensional textual and acoustic representations. For the emotion recognition tasks, the concatenated representation is fully connected to a single output unit by a task-specific linear layer with a 0.1 dropout rate. For the stress detection task, two output units are used to predict the logits for stress and non-stress labels. A softmax layer is used to compute probabilities and training loss. Note that in related work \cite{jaiswal2020muse,bara2020deep}, the ``late fusion'' stands for an ensemble method, while {\sc MUSER} solves the task with a single model.

\subsection{{\sc MUSER}: Multi-task Learning}
\subsubsection{Weight Sharing}
We directly share all the trainable parameters (hard sharing) except the task-specific output layers. For each batch of training samples, only one task is assigned, and one step of back-propagation is performed according to the task objective function with the task-specific output layer plugged in.

\subsubsection{Sampling Strategy}
In each epoch of multi-task training, different amounts of training data are sampled from both the auxiliary task of activation/valence regression and the main task of stress classification. We explore both uniform sampling and dynamic sampling strategy to adaptively decide the mixing ratio of the multiple tasks in each epoch.

\begin{table*}[h]
  \begin{center}
    \caption{Comparison with look-alike methods for multi-task learning.}
    \label{dynamic}
    \scalebox{0.95}{
    \begin{tabular}{l|l|l|c}
    \hline 
    Method & Type & Based on & History \\
    \hline
    \cite{liu2019end} & Loss Balancing & Ratios of training loss & 1\\
    \cite{gottumukkala2020dynamic} & Dynamic Sampling & Validation metrics & 1\\\hline
    MUSER & Dynamic Sampling & Ratios of validation metrics & $n\geq5$\\\hline
    
    \end{tabular}
    }
  \end{center}
\end{table*}

\textbf{Uniform Sampling.} In our conditions, the number of training samples in the main task and the auxiliary tasks are approximately on the same scale. Therefore, an intuitive method is to switch between the tasks with uniform sampling: for each batch, we first decide which task to train with an equal chance, and then randomly select 32 (the batch size) samples; the batch is trained with the corresponding supervision signals (either emotion scores or stress labels) from the selected task.

\textbf{Dynamic Sampling.} Having an equal number of samples for each task in each epoch is not the most efficient way for multi-task training because it is not searching for the most informative task during each epoch. It is more intuitive that when one task reaches a bottleneck, more samples from the other tasks should be selected instead.

Motivated by this idea, we propose to dynamically select the task for each batch according to the model's speed of learning each task. After each training epoch, the sampling distribution is updated based on the model's current and historical performance on each task on the validation set. Specifically, for activation and valence tasks, we compute the ratio of the average rooted mean square error (RMSE) score on the past $n$ epochs to the RMSE score in the current epoch. The ratios are noted as $r_a$ and $r_v$, respectively. For the stress task, we compute the ratio of the accuracy in the current epoch to the average of the past $n$ epochs, noted as $r_s$. The history length $n$ is picked by hand. The sampling distribution for the next epoch is then computed as:
\begin{equation}
    p_a, p_v, p_s = softmax([r_a / \rho, r_v / \rho, r_s / \rho]),
\end{equation}
where $\rho$ is the temperature coefficient set to 0.1.

We use the ``ratios to history'' instead of the validation scores themselves to compute the distribution because this makes different metrics comparable to each other, and it is a good estimation of which task is the model currently learning the fastest. We name this strategy a ``speed-based'' dynamic sampling. The sampling procedure is shown in Figure \ref{fig_dynamic}, and a comparison to look-alike multi-task learning methods is included in Table \ref{dynamic}.


\begin{table*}[h]
  \begin{center}
    \caption{Results: stress detection with single modality.}
    \label{single-modal-stress}
    \scalebox{0.95}{
    \begin{tabular}{l|c|c|c|c}
    \hline 
    Models & Accuracy & Precision & Recall & F-score\\
    \hline
    MLP+LIWC \cite{jaiswal2020muse} & 0.60 & 0.74 & 0.61 & 0.67\\
    MLP+Opensmile \cite{jaiswal2020muse} & 0.67 & 0.70 & 0.69 & 0.69\\\hline
    {\sc MUSER} Textual Encoder & 0.69 & 0.77 & 0.74 & 0.75 \\
    {\sc MUSER} Acoustic Encoder & 0.79 & 0.80 & 0.79 & 0.80 \\\hline
    \end{tabular}
    }
  \end{center}
\end{table*}

\begin{table*}[t]
  \begin{center}
    \caption{Stress detection results with multiple modalities, with or without pre-training.}
    \label{multimodal-single-task}
    \scalebox{0.95}{
    \begin{tabular}{l|c|c|c|c}
    \hline 
    Method & Accuracy & Precision & Recall & F-score\\
    \hline
    LIWC+Audio \cite{jaiswal2020muse} & 0.60 & 0.74 & 0.61 & 0.67 \\
    A-modal GRU \cite{bara2020deep} & 0.573 & 0.582 & 0.557 & 0.569 \\\hline
    {\sc MUSER} from scratch & 0.821 & 0.834 & 0.828 & 0.831 \\
    Activation-100 & \textbf{0.842} & \textbf{0.866} & 0.832 & \textbf{0.849} \\
    Valence-100 & 0.823 & 0.830 & \textbf{0.839} & 0.834\\\hline
    Activation-valence-stress & 0.819 & 0.841 & 0.813 & 0.827\\
    Valence-activation-stress & 0.828 & 0.854 & 0.817 & 0.835\\\hline
    \end{tabular}
    }
  \end{center}
\end{table*}

\begin{table*}[t]
  \begin{center}
    \caption{Stress detection results with multi-task learning}
    \label{multimodal-multi-task}
    \scalebox{0.95}{
    \begin{tabular}{l|c|c|c|c}
    \hline 
    Strategy & Accuracy & Precision & Recall & F-score\\
    \hline
    Uniform sampling with activation & 0.832 & 0.833 & 0.857 & 0.845 \\
    Uniform sampling with valence & 0.823 & 0.837 & 0.828 & 0.832\\
    Uniform sampling with activation \& valence & 0.846 & 0.862 & 0.846 & 0.854 \\\hline
    \cite{liu2019end} - modified & 0.842 & 0.856 & 0.846 & 0.851 \\\hline
    Speed-based sampling (history=5) & 0.854 & 0.861 & \textbf{0.864} & 0.863 \\
    Speed-based sampling (history=10) & \textbf{0.856} & \textbf{0.867} & 0.861 & \textbf{0.864} \\\hline
    
    \end{tabular}
    }
  \end{center}
\end{table*}

\begin{table*}[t]
  \begin{center}
    \caption{Stress detection results with multi-task learning with OMG-Emotions}
    \label{multimodal-multi-omg}
    \scalebox{0.95}{
    \begin{tabular}{l|c|c|c|c}
    \hline 
    Strategy & Accuracy & Precision & Recall & F-score\\
    \hline
    Uniform sampling with activation \& valence & 0.844 & 0.867  & 0.835  &  0.850 \\\hline
    \cite{liu2019end} - modified & 0.836 & 0.820 & 0.886 & 0.852 \\\hline
    Speed-based sampling (history=5) & \textbf{0.850} & \textbf{0.871} & \textbf{0.842} & \textbf{0.856} \\\hline
    
    \end{tabular}
    }
  \end{center}
\end{table*}
 
\section{Experiments}
\subsection{Settings}
We use an AdamW \cite{loshchilov2018decoupled} optimizer with an initial learning rate of 3e-4 for all our multimodal and multi-task experiments. In each epoch, we repeatedly sample data with a batch-size of 32 from the main task or the auxiliary tasks, and apply one-step back-propagation for each batch, until the total selected number reaches the size of the MuSE training set. Gradients are clipped to have a maximum norm of 1.0. The history length $n$ in speed-based dynamic sampling is chosen from \{1, 5, 10\} according to the performance on the validation set. We warm up the dynamic sampling by applying uniform sampling for the first $n$ epochs. The maximum epoch number is typically set to 1000, while the training process is controlled by early stopping. For the Transformer textual encoder, we limit the maximum sequence length to be 128. The evaluation metrics include overall accuracy, as well as the precision, recall, and f-score for the ``stressed'' class. 

    

\subsection{Unimodal Results}
For unimodal experiments, we use the textual encoder or the acoustic encoder independently to compute representations before regression or classification. For the Transformer textual encoder, we use a learning rate of 2e-5; for the MLP acoustic model, we use a learning rate of 5e-4. These learning rates are separately fine-tuned on each unimodal task. Other hyperparameters of the models are kept the same as the multimodal structure.

Table \ref{single-modal-stress} shows the stress detection results with single modalities. Our Transformer encoder outperforms the baseline textual model because of its capability to discover syntactic-level long distance relationships in natural language and the external linguistic knowledge from the advanced BERT pre-training; our acoustic model also improves beyond the baseline results, potentially because we used a more up-to-date version of eGeMaps configuration and a fine-tuned learning rate.

\subsection{Multimodal Results}
To jointly train with both the textual and acoustic features, we use the multimodal fusion model introduced in Section \ref{4.3} as a basic architecture.

\subsubsection{Pre-training}
Our {\sc MUSER} model is trained from scratch to set up a single-task baseline for multimodal stress detection. Besides, a potential alternative to multi-task learning is pre-training on the auxiliary tasks and fine-tuning on the main task. For a complete comparison, we set up several strategies for pre-training. All the pre-training methods use the internal auxiliary task of MuSE. The compared methods are as follows:
\\\textbf{Activation-100}: pre-train for 100 epochs with the activation annotations, then switch to the main task of stress detection. \\\textbf{Valence-100}: pre-train for 100 epochs with the valence annotations, then switch to the main task of stress detection.\\\textbf{Activation-valence-stress}: pre-train for 100 epochs on the activation task, then 100 epochs on the valence task, and switch to stress detection.\\\textbf{Valence-activation-stress}: pre-train for 100 epochs on the valence task, then 100 epochs on the activation task, and switch to stress detection.

The results are presented in Table \ref{multimodal-single-task}. Among the pre-training and fine-tuning results, Activation-100 shows the most significant improvement. The second-best score is the valence-activation-stress order. Thus, we can conclude that activation is the better auxiliary task under this paradigm. Additionally, using only one auxiliary task is always better than using two of them; this is because when the model learns from the second auxiliary task, it ``forgets'' the knowledge from the previous task because it lacks a memory mechanism to look back \cite{hayes2020remind}. 

Pre-training on the emotion recognition tasks using either activation or valence improves stress detection because the model is equipped with the capabilities to encode the features and predict emotions before the training of stress detection task starts.

\subsubsection{Multi-task Learning on MuSE}
For multi-task learning, we compare two sampling strategies: uniform sampling and our proposed speed-based dynamic sampling. We also implement and modify the loss-based weight balancing method proposed by \cite{liu2019end} to adjust the mixing ratios in dynamic sampling instead, and compare it with our methods. The results using the internal MuSE emotion recognition as an auxiliary task are shown in Table \ref{multimodal-multi-task}. 

Comparing the uniform sampling results with Table \ref{multimodal-single-task}, we conclude that using any auxiliary task is better than training from scratch. However, multi-task training with the activation and valence tasks together is better than using them separately. This is different from the observations in Table \ref{multimodal-single-task} and can be explained by the differences in the training procedure: in multi-task learning, the model looks back-and-forth into each task in each epoch, making it able to memorize the shared knowledge from all the tasks. Additionally, when the model is optimized for the two emotion tasks at the same time, the lower-level representation becomes more general and informative because it is frequently plugged with different task-specific layers. 

Comparing the results of using a single auxiliary task of activation vs. valence, activation leads to  better results as compared to valence, which is in agreement with Table \ref{multimodal-single-task}. This is further supported by the analyses in Tables \ref{stress-language} and \ref{uniq}: given the lower unique indicator count of the activation task, as well as the fact that the pre-training and multi-task learning results are all compatible, we can conclude that for stress detection, the nature of the activation dimension of emotion is closer and more helpful than the valence dimension. This potentially suggests that stress has a major effect on whether people feel excited (activation), but a minor effect on their opinion toward events and objects (valence).

We test our speed-based dynamic sampling algorithm using activation and valence together as auxiliary tasks and it yields promising results with history set to 5 and 10. It significantly outperforms both the uniform sampling and our implementation of the loss-based strategy \cite{liu2019end} (t-test, $p < 0.05$), achieving state-of-the-art scores on MuSE stress detection task with one single model and only two modalities.

Our model works the best with a history length between 5 and 10. If the history is too short, the model takes longer to converge and has unstable performance, while if the history is too long, it fails to capture the dynamics.

\subsubsection{Generalization}
In real-world applications, stress detection data does not necessarily have emotion labels. However, because of the intrinsic inter-dependence between emotion and stress, any existing dataset with emotion labels can potentially serve as an external auxiliary task for stress detection. However, this requires our model and multi-task training algorithm to generalize beyond the internal MuSE emotion tasks. We test our model on OMG-Emotions as an example of external emotion datasets.

Table \ref{multimodal-multi-omg} shows results on MuSE stress detection using OMG-Emotion as an auxiliary task. Comparing to Table \ref{multimodal-single-task}, although the source and content of OMG-Emotions are different from MuSE, multi-task learning still outperforms single-task learning and pre-training (t-test, $p < 0.05$). This reveals that the connection between stress and emotion widely exists, and our multi-task learning method works in general cases. 

Additionally, Table \ref{multimodal-multi-omg} suggests that while using an external emotion dataset, our speed-based sampling method still outperforms uniform sampling, as well as our implementation of loss-based dynamic sampling \cite{liu2019end}. This supports the robustness and effectiveness of our speed-based strategy.

\section{Conclusions}
In this work, we uncovered the connections and differences between stress detection and emotion recognition using textual and acoustic features, and proposed to use emotion recognition as an auxiliary task for stress detection. We proposed {\sc MUSER}: a Transformer-based model structure, together with a novel speed-based dynamic sampling strategy for multi-task learning. Experimental results support the inter-dependence of stress and emotion (activation/valence), and proves the effectiveness and robustness of our methods. {\sc MUSER} achieved state-of-the-art results on the MuSE stress detection task both when internal (MuSE) and when external (OMG-Emotions) emotion data and annotations were used.

Our code is publicly available at \url{https://lit.eecs.umich.edu/downloads.html#MUSER}

\section*{Acknowledgements}
We would like to thank Cristian-Paul Bara and Mimansa Jaiswal for their helpful discussion on the data processing and features of MuSE dataset. We appreciate the insightful comments from all the reviewers and program committee.

This material is based in part on work supported by the Toyota Research Institute, the Precision Health initiative at the University of Michigan, the National Science Foundation (grant \#1815291), and the John Templeton Foundation (grant \#61156). Any opinions, findings, conclusions, or recommendations in this material are those of the authors and do not necessarily reflect the views of the Toyota Research Institute, Precision Health initiative, the National Science Foundation, or the John Templeton Foundation. 
\bibliography{anthology,references}

\begin{thebibliography}{48}
\expandafter\ifx\csname natexlab\endcsname\relax\def\natexlab#1{#1}\fi

\bibitem[{Adolphs(2003)}]{adolphs2003human}
Ralph Adolphs. 2003.
\newblock Is the human amygdala specialized for processing social information?
\newblock \emph{Annals of the New York Academy of Sciences}, 985(1):326--340.

\bibitem[{Aigrain et~al.(2016)Aigrain, Spodenkiewicz, Dubuiss, Detyniecki,
  Cohen, and Chetouani}]{aigrain2016multimodal}
Jonathan Aigrain, Michel Spodenkiewicz, S{\'e}verine Dubuiss, Marcin
  Detyniecki, David Cohen, and Mohamed Chetouani. 2016.
\newblock Multimodal stress detection from multiple assessments.
\newblock \emph{IEEE Transactions on Affective Computing}, 9(4):491--506.

\bibitem[{Alberdi et~al.(2016)Alberdi, Aztiria, and
  Basarab}]{alberdi2016towards}
Ane Alberdi, Asier Aztiria, and Adrian Basarab. 2016.
\newblock Towards an automatic early stress recognition system for office
  environments based on multimodal measurements: A review.
\newblock \emph{Journal of biomedical informatics}, 59:49--75.

\bibitem[{Bara et~al.(2020)Bara, Papakostas, and Mihalcea}]{bara2020deep}
Cristian~Paul Bara, Michalis Papakostas, and Rada Mihalcea. 2020.
\newblock A deep learning approach towards multimodal stress detection.
\newblock In \emph{Proceedings of the AAAI-20 Workshop on Affective Content
  Analysis, New York, USA, AAAI}.

\bibitem[{Barros et~al.(2018)Barros, Churamani, Lakomkin, Siqueira, Sutherland,
  and Wermter}]{barros2018omg}
Pablo Barros, Nikhil Churamani, Egor Lakomkin, Henrique Siqueira, Alexander
  Sutherland, and Stefan Wermter. 2018.
\newblock The omg-emotion behavior dataset.
\newblock In \emph{2018 International Joint Conference on Neural Networks
  (IJCNN)}, pages 1--7. IEEE.

\bibitem[{Botelho et~al.(2019)Botelho, Trancoso, Abad, and
  Paiva}]{botelho2019speech}
M~Catarina Botelho, Isabel Trancoso, Alberto Abad, and Teresa Paiva. 2019.
\newblock Speech as a biomarker for obstructive sleep apnea detection.
\newblock In \emph{ICASSP 2019-2019 IEEE International Conference on Acoustics,
  Speech and Signal Processing (ICASSP)}, pages 5851--5855. IEEE.

\bibitem[{Bradley and Lang(1994)}]{bradley1994measuring}
Margaret~M Bradley and Peter~J Lang. 1994.
\newblock Measuring emotion: the self-assessment manikin and the semantic
  differential.
\newblock \emph{Journal of behavior therapy and experimental psychiatry},
  25(1):49--59.

\bibitem[{Busso et~al.(2008)Busso, Bulut, Lee, Kazemzadeh, Mower, Kim, Chang,
  Lee, and Narayanan}]{busso2008iemocap}
Carlos Busso, Murtaza Bulut, Chi-Chun Lee, Abe Kazemzadeh, Emily Mower, Samuel
  Kim, Jeannette~N Chang, Sungbok Lee, and Shrikanth~S Narayanan. 2008.
\newblock Iemocap: Interactive emotional dyadic motion capture database.
\newblock \emph{Language resources and evaluation}, 42(4):335.

\bibitem[{Busso et~al.(2016)Busso, Parthasarathy, Burmania, AbdelWahab,
  Sadoughi, and Provost}]{busso2016msp}
Carlos Busso, Srinivas Parthasarathy, Alec Burmania, Mohammed AbdelWahab,
  Najmeh Sadoughi, and Emily~Mower Provost. 2016.
\newblock Msp-improv: An acted corpus of dyadic interactions to study emotion
  perception.
\newblock \emph{IEEE Transactions on Affective Computing}, 8(1):67--80.

\bibitem[{Chen et~al.(2017)Chen, Zhao, Ye, Zhang, and Zou}]{chen2017detecting}
Lan-lan Chen, Yu~Zhao, Peng-fei Ye, Jian Zhang, and Jun-zhong Zou. 2017.
\newblock Detecting driving stress in physiological signals based on multimodal
  feature analysis and kernel classifiers.
\newblock \emph{Expert Systems with Applications}, 85:279--291.

\bibitem[{Chen et~al.(2018{\natexlab{a}})Chen, Hsu, Kuo, Ku
  et~al.}]{chen2018emotionlines}
Sheng-Yeh Chen, Chao-Chun Hsu, Chuan-Chun Kuo, Lun-Wei Ku, et~al.
  2018{\natexlab{a}}.
\newblock Emotionlines: An emotion corpus of multi-party conversations.
\newblock \emph{arXiv preprint arXiv:1802.08379}.

\bibitem[{Chen et~al.(2018{\natexlab{b}})Chen, Badrinarayanan, Lee, and
  Rabinovich}]{chen2018gradnorm}
Zhao Chen, Vijay Badrinarayanan, Chen-Yu Lee, and Andrew Rabinovich.
  2018{\natexlab{b}}.
\newblock Gradnorm: Gradient normalization for adaptive loss balancing in deep
  multitask networks.
\newblock In \emph{International Conference on Machine Learning}, pages
  794--803. PMLR.

\bibitem[{Cohen et~al.(1991)Cohen, Tyrrell, and Smith}]{cohen1991psychological}
Sheldon Cohen, David~AJ Tyrrell, and Andrew~P Smith. 1991.
\newblock Psychological stress and susceptibility to the common cold.
\newblock \emph{New England journal of medicine}, 325(9):606--612.

\bibitem[{Devlin et~al.(2019)Devlin, Chang, Lee, and
  Toutanova}]{devlin2019bert}
Jacob Devlin, Ming-Wei Chang, Kenton Lee, and Kristina Toutanova. 2019.
\newblock Bert: Pre-training of deep bidirectional transformers for language
  understanding.
\newblock In \emph{Proceedings of the 2019 Conference of the North American
  Chapter of the Association for Computational Linguistics: Human Language
  Technologies, Volume 1 (Long and Short Papers)}, pages 4171--4186.

\bibitem[{Dobson and Smith(2000)}]{dobson2000stress}
Hilary Dobson and RF~Smith. 2000.
\newblock What is stress, and how does it affect reproduction?
\newblock \emph{Animal reproduction science}, 60:743--752.

\bibitem[{Eyben et~al.(2015)Eyben, Scherer, Schuller, Sundberg, Andr{\'e},
  Busso, Devillers, Epps, Laukka, Narayanan et~al.}]{eyben2015geneva}
Florian Eyben, Klaus~R Scherer, Bj{\"o}rn~W Schuller, Johan Sundberg, Elisabeth
  Andr{\'e}, Carlos Busso, Laurence~Y Devillers, Julien Epps, Petri Laukka,
  Shrikanth~S Narayanan, et~al. 2015.
\newblock The geneva minimalistic acoustic parameter set (gemaps) for voice
  research and affective computing.
\newblock \emph{IEEE transactions on affective computing}, 7(2):190--202.

\bibitem[{Eyben et~al.(2010)Eyben, W{\"o}llmer, and
  Schuller}]{eyben2010opensmile}
Florian Eyben, Martin W{\"o}llmer, and Bj{\"o}rn Schuller. 2010.
\newblock Opensmile: the munich versatile and fast open-source audio feature
  extractor.
\newblock In \emph{Proceedings of the 18th ACM international conference on
  Multimedia}, pages 1459--1462.

\bibitem[{Gao et~al.(2014)Gao, Y{\"u}ce, and Thiran}]{gao2014detecting}
Hua Gao, Anil Y{\"u}ce, and Jean-Philippe Thiran. 2014.
\newblock Detecting emotional stress from facial expressions for driving
  safety.
\newblock In \emph{2014 IEEE International Conference on Image Processing
  (ICIP)}, pages 5961--5965. IEEE.

\bibitem[{Gong et~al.(2019)Gong, Lee, Stephenson, Renduchintala, Padhy,
  Ndirango, Keskin, and Elibol}]{gong2019comparison}
Ting Gong, Tyler Lee, Cory Stephenson, Venkata Renduchintala, Suchismita Padhy,
  Anthony Ndirango, Gokce Keskin, and Oguz~H Elibol. 2019.
\newblock A comparison of loss weighting strategies for multi task learning in
  deep neural networks.
\newblock \emph{IEEE Access}, 7:141627--141632.

\bibitem[{Gottumukkala et~al.(2020)Gottumukkala, Dua, Singh, and
  Gardner}]{gottumukkala2020dynamic}
Ananth Gottumukkala, Dheeru Dua, Sameer Singh, and Matt Gardner. 2020.
\newblock Dynamic sampling strategies for multi-task reading comprehension.
\newblock In \emph{Proceedings of the 58th Annual Meeting of the Association
  for Computational Linguistics}, pages 920--924.

\bibitem[{Guo et~al.(2018{\natexlab{a}})Guo, Pasunuru, and
  Bansal}]{guo2018dynamic}
Han Guo, Ramakanth Pasunuru, and Mohit Bansal. 2018{\natexlab{a}}.
\newblock Dynamic multi-level multi-task learning for sentence simplification.
\newblock In \emph{Proceedings of the 27th International Conference on
  Computational Linguistics}, pages 462--476.

\bibitem[{Guo et~al.(2018{\natexlab{b}})Guo, Haque, Huang, Yeung, and
  Fei-Fei}]{feifei2018dynamic}
Michelle Guo, Albert Haque, De-An Huang, Serena Yeung, and Li~Fei-Fei.
  2018{\natexlab{b}}.
\newblock Dynamic task prioritization for multitask learning.
\newblock In \emph{Proceedings of the European Conference on Computer Vision
  (ECCV)}, pages 270--287.

\bibitem[{Hayes et~al.(2020)Hayes, Kafle, Shrestha, Acharya, and
  Kanan}]{hayes2020remind}
Tyler~L Hayes, Kushal Kafle, Robik Shrestha, Manoj Acharya, and Christopher
  Kanan. 2020.
\newblock Remind your neural network to prevent catastrophic forgetting.
\newblock In \emph{European Conference on Computer Vision}, pages 466--483.
  Springer.

\bibitem[{Horvath(1982)}]{horvath1982detecting}
Frank Horvath. 1982.
\newblock Detecting deception: the promise and the reality of voice stress
  analysis.
\newblock \emph{Journal of Forensic Science}, 27(2):340--351.

\bibitem[{Jaiswal et~al.(2019)Jaiswal, Aldeneh, Bara, Luo, Burzo, Mihalcea, and
  Provost}]{jaiswal2019muse}
Mimansa Jaiswal, Zakaria Aldeneh, Cristian-Paul Bara, Yuanhang Luo, Mihai
  Burzo, Rada Mihalcea, and Emily~Mower Provost. 2019.
\newblock Muse-ing on the impact of utterance ordering on crowdsourced emotion
  annotations.
\newblock In \emph{ICASSP 2019-2019 IEEE International Conference on Acoustics,
  Speech and Signal Processing (ICASSP)}, pages 7415--7419. IEEE.

\bibitem[{Jaiswal et~al.(2020)Jaiswal, Bara, Luo, Burzo, Mihalcea, and
  Provost}]{jaiswal2020muse}
Mimansa Jaiswal, Cristian-Paul Bara, Yuanhang Luo, Mihai Burzo, Rada Mihalcea,
  and Emily~Mower Provost. 2020.
\newblock Muse: a multimodal dataset of stressed emotion.
\newblock In \emph{Proceedings of The 12th Language Resources and Evaluation
  Conference}, pages 1499--1510.

\bibitem[{Jaques et~al.(2016)Jaques, Taylor, Nosakhare, Sano, and
  Picard}]{jaques2016multi}
Natasha Jaques, Sara Taylor, Ehimwenma Nosakhare, Akane Sano, and Rosalind
  Picard. 2016.
\newblock Multi-task learning for predicting health, stress, and happiness.
\newblock In \emph{NIPS Workshop on Machine Learning for Healthcare}.

\bibitem[{Kendall et~al.(2018)Kendall, Gal, and Cipolla}]{kendall2018multi}
Alex Kendall, Yarin Gal, and Roberto Cipolla. 2018.
\newblock Multi-task learning using uncertainty to weigh losses for scene
  geometry and semantics.
\newblock In \emph{Proceedings of the IEEE conference on computer vision and
  pattern recognition}, pages 7482--7491.

\bibitem[{Lample et~al.(2017)Lample, Conneau, Denoyer, and
  Ranzato}]{lample2017unsupervised}
Guillaume Lample, Alexis Conneau, Ludovic Denoyer, and Marc'Aurelio Ranzato.
  2017.
\newblock Unsupervised machine translation using monolingual corpora only.
\newblock \emph{arXiv preprint arXiv:1711.00043}.

\bibitem[{Lane et~al.(2015)Lane, Georgiev, and Qendro}]{lane2015deepear}
Nicholas~D Lane, Petko Georgiev, and Lorena Qendro. 2015.
\newblock Deepear: robust smartphone audio sensing in unconstrained acoustic
  environments using deep learning.
\newblock In \emph{Proceedings of the 2015 ACM International Joint Conference
  on Pervasive and Ubiquitous Computing}, pages 283--294.

\bibitem[{Lech and He(2014)}]{lech2014stress}
Margaret Lech and Ling He. 2014.
\newblock Stress and emotion recognition using acoustic speech analysis.
\newblock In \emph{Mental Health Informatics}, pages 163--184. Springer.

\bibitem[{Lin et~al.(2014)Lin, Jia, Guo, Xue, Li, Huang, Cai, and
  Feng}]{lin2014user}
Huijie Lin, Jia Jia, Quan Guo, Yuanyuan Xue, Qi~Li, Jie Huang, Lianhong Cai,
  and Ling Feng. 2014.
\newblock User-level psychological stress detection from social media using
  deep neural network.
\newblock In \emph{Proceedings of the 22nd ACM international conference on
  Multimedia}, pages 507--516.

\bibitem[{Liu et~al.(2019)Liu, Johns, and Davison}]{liu2019end}
Shikun Liu, Edward Johns, and Andrew~J Davison. 2019.
\newblock End-to-end multi-task learning with attention.
\newblock In \emph{Proceedings of the IEEE Conference on Computer Vision and
  Pattern Recognition}, pages 1871--1880.

\bibitem[{Loshchilov and Hutter(2018)}]{loshchilov2018decoupled}
Ilya Loshchilov and Frank Hutter. 2018.
\newblock Decoupled weight decay regularization.
\newblock In \emph{International Conference on Learning Representations}.

\bibitem[{McKeown et~al.(2011)McKeown, Valstar, Cowie, Pantic, and
  Schroder}]{mckeown2011semaine}
Gary McKeown, Michel Valstar, Roddy Cowie, Maja Pantic, and Marc Schroder.
  2011.
\newblock The semaine database: Annotated multimodal records of emotionally
  colored conversations between a person and a limited agent.
\newblock \emph{IEEE transactions on affective computing}, 3(1):5--17.

\bibitem[{Muthukumar and
  Nachiappan(2013)}]{muthukumar2013phosphatidylethanolamine}
Kannan Muthukumar and Vasanthi Nachiappan. 2013.
\newblock Phosphatidylethanolamine from phosphatidylserine decarboxylase2 is
  essential for autophagy under cadmium stress in saccharomyces cerevisiae.
\newblock \emph{Cell biochemistry and biophysics}, 67(3):1353--1363.

\bibitem[{Pasunuru and Bansal(2017)}]{pasunuru2017multi}
Ramakanth Pasunuru and Mohit Bansal. 2017.
\newblock Multi-task video captioning with video and entailment generation.
\newblock In \emph{Proceedings of the 55th Annual Meeting of the Association
  for Computational Linguistics (Volume 1: Long Papers)}, pages 1273--1283.

\bibitem[{Paulmann et~al.(2016)Paulmann, Furnes, B{\o}kenes, and
  Cozzolino}]{paulmann2016psychological}
Silke Paulmann, Desire Furnes, Anne~Ming B{\o}kenes, and Philip~J Cozzolino.
  2016.
\newblock How psychological stress affects emotional prosody.
\newblock \emph{Plos one}, 11(11):e0165022.

\bibitem[{Pennebaker et~al.(2001)Pennebaker, Francis, and
  Booth}]{pennebaker2001linguistic}
James~W Pennebaker, Martha~E Francis, and Roger~J Booth. 2001.
\newblock Linguistic inquiry and word count: Liwc 2001.
\newblock \emph{Mahway: Lawrence Erlbaum Associates}, 71(2001):2001.

\bibitem[{Pruessner et~al.(2008)Pruessner, Dedovic, Khalili-Mahani, Engert,
  Pruessner, Buss, Renwick, Dagher, Meaney, and
  Lupien}]{pruessner2008deactivation}
Jens~C Pruessner, Katarina Dedovic, Najmeh Khalili-Mahani, Veronika Engert,
  Marita Pruessner, Claudia Buss, Robert Renwick, Alain Dagher, Michael~J
  Meaney, and Sonia Lupien. 2008.
\newblock Deactivation of the limbic system during acute psychosocial stress:
  evidence from positron emission tomography and functional magnetic resonance
  imaging studies.
\newblock \emph{Biological psychiatry}, 63(2):234--240.

\bibitem[{Taylor and Stanton(2007)}]{taylor2007coping}
Shelley~E Taylor and Annette~L Stanton. 2007.
\newblock Coping resources, coping processes, and mental health.
\newblock \emph{Annu. Rev. Clin. Psychol.}, 3:377--401.

\bibitem[{Tull et~al.(2007)Tull, Barrett, McMillan, and
  Roemer}]{tull2007preliminary}
Matthew~T Tull, Heidi~M Barrett, Elaine~S McMillan, and Lizabeth Roemer. 2007.
\newblock A preliminary investigation of the relationship between emotion
  regulation difficulties and posttraumatic stress symptoms.
\newblock \emph{Behavior Therapy}, 38(3):303--313.

\bibitem[{Vaswani et~al.(2017)Vaswani, Shazeer, Parmar, Uszkoreit, Jones,
  Gomez, Kaiser, and Polosukhin}]{vaswani2017attention}
Ashish Vaswani, Noam Shazeer, Niki Parmar, Jakob Uszkoreit, Llion Jones,
  Aidan~N Gomez, {\L}ukasz Kaiser, and Illia Polosukhin. 2017.
\newblock Attention is all you need.
\newblock In \emph{Advances in neural information processing systems}, pages
  5998--6008.

\bibitem[{Wang and Saudino(2011)}]{wang2011emotion}
Manjie Wang and Kimberly~J Saudino. 2011.
\newblock Emotion regulation and stress.
\newblock \emph{Journal of Adult Development}, 18(2):95--103.

\bibitem[{Yao et~al.(2020)Yao, P{\'e}rez-Rosas, Abouelenien, and
  Burzo}]{yao2020morse}
Yiqun Yao, Ver{\'o}nica P{\'e}rez-Rosas, Mohamed Abouelenien, and Mihai Burzo.
  2020.
\newblock Morse: Multimodal sentiment analysis for real-life settings.
\newblock In \emph{Proceedings of the 2020 International Conference on
  Multimodal Interaction}, pages 387--396.

\bibitem[{Yao et~al.(2019)Yao, Xu, and Xu}]{yao2019world}
Yiqun Yao, Jiaming Xu, and Bo~Xu. 2019.
\newblock The world in my mind: Visual dialog with adversarial multi-modal
  feature encoding.
\newblock In \emph{Proceedings of the 2019 Conference of the North American
  Chapter of the Association for Computational Linguistics: Human Language
  Technologies, Volume 1 (Long and Short Papers)}, pages 2588--2598.

\bibitem[{Zadeh et~al.(2016)Zadeh, Zellers, Pincus, and
  Morency}]{zadeh2016mosi}
Amir Zadeh, Rowan Zellers, Eli Pincus, and Louis-Philippe Morency. 2016.
\newblock Mosi: multimodal corpus of sentiment intensity and subjectivity
  analysis in online opinion videos.
\newblock \emph{arXiv preprint arXiv:1606.06259}.

\bibitem[{Zadeh et~al.(2018)Zadeh, Liang, Poria, Cambria, and
  Morency}]{zadeh2018multimodal}
AmirAli~Bagher Zadeh, Paul~Pu Liang, Soujanya Poria, Erik Cambria, and
  Louis-Philippe Morency. 2018.
\newblock Multimodal language analysis in the wild: Cmu-mosei dataset and
  interpretable dynamic fusion graph.
\newblock In \emph{Proceedings of the 56th Annual Meeting of the Association
  for Computational Linguistics (Volume 1: Long Papers)}, pages 2236--2246.

\end{thebibliography}
\bibliographystyle{acl_natbib}




\end{document}